\pdfoutput=1
\documentclass[twoside,11pt,dvipsnames]{fairmeta}
\usepackage{times}
\usepackage{latexsym}

\usepackage[T1]{fontenc}
\usepackage[utf8]{inputenc}
\usepackage{booktabs}
\usepackage{microtype}

\usepackage{inconsolata}

\usepackage{graphicx}
\usepackage{color}
\usepackage{colortbl}

\usepackage{comment}
\affiliation[]{FAIR at Meta}

\date{December 12, 2024}
\correspondence{Marta R. Costa-jussà at \email{costajussa@meta.com}}

\title{{Y-NQ:}\\[7pt]\Large  English-Yor\`{u}b\'{a} Evaluation dataset for Open-Book Reading Comprehension and Text Generation}

\author[]{Marta R. Costa-jussà, Joy Chen, Ifeoluwanimi Adebara, Joe Chuang, Christophe Ropers, Eduardo Sánchez}
\abstract{The purpose of this work is to share an English-Yor\`{u}b\'{a} evaluation dataset for open-book reading comprehension and text generation to assess the performance of models both in a high- and a low-resource language. The dataset contains 358 questions and answers on 338 English documents and 208 Yor\`{u}b\'{a} documents. The average document length is $\approx$ 10k words for English and 430 words for Yor\`{u}b\'{a}. %The average question length is around 9 words for both languages and the average answer is approximately 114 words for English and 33 for Yor\`{u}b\'{a}. 
Experiments show a consistent disparity in performance between the two languages, with Yor\`{u}b\'{a} falling behind English for automatic metrics even if documents are much shorter for this language. %However, human evaluation tells a slightly different story and Yor\`{u}b\'{a} is slightly better than English. 
For a small set of documents with comparable length, performance of Yor\`{u}b\'{a} drops by $x$2.5 times. %, but human evaluation scores again better for Yor\`{u}b\'{a} . 
When analyzing performance by length, we observe that Yor\`{u}b\'{a} decreases performance dramatically for documents that reach 1500 words while English performance is barely affected at that length. Our dataset opens the door to showcasing if English LLM reading comprehension capabilities extend to Yor\`{u}b\'{a}, which for the evaluated LLMs is not the case.}

\begin{document}
\maketitle

\section{Introduction}

%{\color{red}  the purpose of the task is to show if the model answers with memorised data. also, even given that the yoruba task is much easier than english, it still falls behind it. so we show how bad llm are in extending capabilities to low resource languages. english performs better because it memorises data or because it is much better on this task}

This study explores the intersection of reading comprehension and text generation, examining how models perform on tasks requiring both in-context understanding (i.e., open-book model, where the model has access to the context document during inference to answer a particular question) and generative text production (i.e. the answer is free-text which has to be compared to a gold standard reference). We aim to investigate the performance of this task in two languages: a high-resource language (English) and a low-resource language (Yor\`{u}b\'{a}).
For this, we introduce Y-NQ (Yor\`{u}b\'{a} Natural Questions) a comprehensive open-book question-answer dataset (Section \ref{sec:dataset}). Y-NQ is sourced from NQ \citep{47761} and provides a complete article context for informed answers and text generation tasks, and parallel documents on the same topic for both high- and low-resource languages. The data set also includes the comparability of the responses in languages. As a result, we are increasing Natural Language Processing (NLP) resources in Yor\`{u}b\'{a} \citep{ahia-etal-2024-voices}. Our data set is benchmarked against state-of-the-art Large Language Models (LLMs). The results and analysis (Section \ref{sec:experiments}) shows that responses in Yor\`{u}b\'{a} are more inaccurate than those in English. %even if the Yor\`{u}b\'{a} documents on which questions are asked are shorter than English ones.

As a by-product of human annotations, we identify inaccuracies in the English-language version of some Wikipedia articles (26 incorrect answers out of 1,566 humanly analyzed questions in the English-language subset of articles), which confirms the existence of accuracy discrepancies across languages for the same Wikipedia topics, thus supporting, for example, the need to better interlink Wikipedia articles across languages \citep{klang-nugues-2016-pairing}. 

%relevant bib: https://aclanthology.org/2021.emnlp-main.493.pdf
%https://arxiv.org/abs/2002.08910

\section{Dataset description}
\label{sec:dataset}

\subsection{Requirements and Background}

The performance of Reading Comprehension (RC) in LLMs has been explored in different settings. At the high level, RC tasks can fall under two main categories: open-book tasks, such as in SQuAD \citep{Rajpurkar2016}, and close-book tasks, such as in TriviaQA \citep{Joshi2017}. Response formats vary across RC tasks as well and include: true/false classification (e.g., BoolQ; \citealp{Clark2019}), multiple-choice questions (e.g., Belebele), span selection (e.g., SQuAD), and text generation (e.g., NQ or TriviaQA).

Since we are interested in exploring the intersection of reading comprehension and text generation covering both a high- and a low-resource language, we can explicitly set our requirements to include for each of the two types of language: (a) long articles ($>$100s words), (b) question-answer pairs with lengthy answers ($>$10s words), and (c) equivalence annotations for cross-lingual answers. Since there are no existing data sets to this effect, we extend existing research by tailoring an established data set to our specific requirements. We justify our choice of data sets and low-resource language selection as explained in the following.

\paragraph{Dataset.} Among the open-book and text generation tasks, one of the largest datasets with multilingual information available is NQ.

\begin{table*}[ht!]
\centering
\small
\begin{tabular}{ll}
\toprule
\textbf{Objective} & Read an article and find a paragraph containing enough information to answer a \\ & specific question. \\ 
\midrule
\textbf{Project Context} & Evaluate accuracy of large language models in finding long contexts and short \\ &  answers; extend Natural Questions dataset to multilingual, non-English centric. \\ 
\midrule
\textbf{Task Components} & \textbullet~ QUESTION: Simple question requesting information or explanation. \\
& \textbullet~ ARTICLE: Numbered paragraphs containing relevant information. \\ 
\midrule
\textbf{Task Steps} & 1. Read QUESTION carefully. \\
& 2. Read ARTICLE paragraphs until sufficient information is found. \\
& 3. Record findings by answering task questions. \\ 
\midrule
\textbf{Additional task steps} & Discard questions that contain the answer in English in the Yor\`{u}b\'{a} document\\
& When possible, add Yor\`{u}b\'{a} questions, translate them into English, and find answers \\ & both in the Yor\`{u}b\'{a} and English documents. \\
\bottomrule
\end{tabular}

\caption{Linguistic guidelines and annotation}
\label{tab:project-overview}
\end{table*}

\paragraph{Low-resource language.} There is a large number of low-resource languages that could be explored here. We prioritize a low-resource language that has overall limited digital resources (in compliance with the definition of low resource), but has a high representation in Wikipedia (on the order of several thousands of entries) and a significant number of speakers (in the order of tens of millions), and makes use of the same script (Latin) as the high-resource language in which results are compared. % it is represented in SONAR with average quality because we are using such tool. 
 One of the languages that complies with all these criteria is Yor\`{u}b\'{a}, in which we can also find works on comprehension of the language in the domain of language exams \citep{aremu2024naijarcmultichoicereadingcomprehension}, based on short passages and multiple choice answers. Another work is the AfriQA dataset \citep{ogundepo-etal-2023-cross} for answering open-retrieval questions, with a primary focus on retrieving correct answers that are answerable on Wikipedia. However, this cannot be used as an open book. %, in-context is not possible, only questions and answers, only close-book.
Finally, Bebebele \citep{bandarkar-etal-2024-belebele} also includes Yor\`{u}b\'{a}, although it uses short passages and multiple choice answers.

\subsection{Dataset creation}

\paragraph{NQ pre-selection.} We looked at 315,203 examples and 231,695 unique English Wikipedia pages from the NQ training and validation datasets.  We filter questions for only those where every long answer is contained in an html tag $<p>$ where $<p>$ is the first identified html tag in the long answer span. This filters out about 25 percent of the questions.

We extracted 2,855 Yor\`{u}b\'{a} Wikipedia pages that are actively associated with the above English pages. We removed documents with fewer than 500 characters, including formatting, and performed multiple cleaning procedures, such as removing html formatting, removing citation notations, and filtering out irrelevant sections in Wikipedia articles (e.g., references, tables). %{\color{red} Joy, can we revise 664? if we sent 664 documents and we discarded 514 documents, how we can have 208 documents??? !!! reply in Nov 30: No sure where we got the 516 number, might due to different iterations. changed the number to 514 in annotation findings section to keep it consistent} 
664 Yor\`{u}b\'{a} documents and 1,566 questions were sent for human annotation.

\paragraph{Pre-annotation effort.} %{\color{red} Joy please check this wording} 
In order to reduce the annotation workload, we automatically pre-selected Yor\`{u}b\'{a} sentences that could be good response candidates by computing a similarity score. If the answer to the question was in agreement with a high similarity score, the annotator would save time by looking through the document and only checking if the match was correct.  %increase the chances of questions being answered in the Yor\`{u}b\'{a} pages, 
We conducted a SONAR embedding similarity \citep{duquenne2023sonarsentencelevelmultimodallanguageagnostic} analysis between Yor\`{u}b\'{a} documents and long English answers. We used the Stopes\footnote{https://github.com/facebookresearch/stopes} sensitizers on all text extracted from $<p>$ elements for both the scraped Yor\`{u}b\'{a} Wikipedia articles downloaded from the previous step and the original NQ Wikipedia pages. We then created SONAR embeddings of each extracted sentence and identified those sentences in the Yor\`{u}b\'{a} pages which were most similar to sentences in the long English answers based on their cosine similarity scores. For a small set of samples, we asked the annotators to examine the entries in a small validation data set to identify a reasonable threshold indicating high similarity between Yor\`{u}b\'{a}/English sentences, which could then be applied to the rest of the data set. The analysis shows a low similarity matching rate, %between Yor\`{u}b\'{a} articles and English answers, 
which is likely due to the low quality and short length of many Yor\`{u}b\'{a} articles and/or SONAR embeddings not being suitable for such a task. Given this low reliability, we abandoned this automatic pre-annotation, which would not reduce annotation efforts.

\paragraph{Annotation guidelines and requirements.}
We designed the annotation guidelines as follows. We provided context on the objective of the task together with the project context and description of the task. The guidelines are summarized in Table \ref{tab:project-overview}.

Finally, beyond the guidelines, we provided additional examples and requested that annotators should be native speakers of the language of the source documents and should have at least CEFR C2 level proficiency in English.

\begin{table}[h!]
\centering
\small
\begin{tabular}{l|l|l}
\hline
&\textsc{Eng} & \textsc{Yor} \\ \hline
\textsc{$\#$Q\&A} & 358 & 358 \\  \hline
\textsc{$\#$docs} & 338 & 208\\ \hline
\textsc{avg. doc len} & 10363 & 430 \\\hline
\textsc{median doc len} & 9272 & 172 \\\hline
\textsc{avg. question len}  & 8.86 & 9.39 \\\hline
\textsc{avg. long answer len}  & 113.80 & 32.89 \\\hline
\end{tabular}
\caption{Dataset Statistics. Length is in words. }
\label{tab:statistics}
\end{table}

\paragraph{Annotator findings.}
We noticed that many articles have a significant amount of English content. Several documents also contained errors, such as incorrect spelling, ungrammatical sentences, and sentences that lacked clarity or meaning. We disregarded such articles and corrected articles that were contaminated with a small amount of English content. We also removed the entries where no answers could be found in the Yor\`{u}b\'{a} articles. %After cleaning, we had {\color{green}\textbf{XX}} documents and {\color{green}\textbf{XX}} questions.

Following the guidelines, the annotators encountered the following: (a) questions with multiple correct answers, for which they annotated each correct answer for the question; (b) questions with correct answers in Yor\`{u}b\'{a}, but incorrect in English, where they annotated the Yor\`{u}b\'{a} appropriately, but flagged the English portion incorrect (there were 26 questions in the category); (c) unclear questions (5 questions) to which no annotations were assigned; (d) answers existing in multiple paragraphs in the document for which they annotated the row with all paragraphs where 

There were 456 Yor\`{u}b\'{a} documents that did not answer the question; therefore, we discarded those. Only eight incorrect English answers from the previous 26 remain in the final dataset, and we did not correct them since the English documents remained the same as in the original NQ.

%{\color{green} [TODO IFE] X documents with more than 50\% in English and with the answer responded in English not in Yor\`{u}b\'{a}. can you add the description and numbers with repairing and adding questions?}

\paragraph{Statistics.}

 Table~\ref{tab:statistics} details the statistics of the data set\footnote{There are two questions that come from the validation NQ dataset, which have two different answers}. Our carefully curated selection contains 208 unique Yor\`{u}b\'{a} Wikipedia documents with an average word count of 430, and 356 unique questions. Only the questions are strictly comparable. %{\color{red}Are topics of documents totally comparable?} 
 English and Yor\`{u}b\'{a} documents are not comparable in number or length, but they are so in topic and domain. The answers are not comparable in length. 
 %The difference in number of documents come from the fact that multiple English topics mapped to the same Yor\`{u}b\'{a} topic and the same English topic has multiple versions of wiki docs in NQ. 
 Notice that English documents outnumber Yor\`{u}b\'{a} documents mainly due to multiple versions of the same English topic counted as different documents, while in Yor\`{u}b\'{a} we selected one version of the document and multiple topics in English that correspond to the same Yor\`{u}b\'{a} topic.
 
 The fact that English documents are longer than those in Yor\`{u}b\'{a} makes the task easier for Yor\`{u}b\'{a}, since documents are significantly shorter within the same topic or domain. We identified a subset of six documents that are strictly comparable in length and topic for English and Yor\`{u}b\'{a}, which allows us to make a fair comparison. Table \ref{tab:dataset-fields} shows the list of fields in Y-NQ and a sample entry.

%Notice that English documents outnumbers Yor\`{u}b\'{a} documents. They mainly due to 1) multiple versions of the same English topic are counted as different documents, while in Yor\`{u}b\'{a} we selected the one version of the document 2) multiple topics in English can map to the same Yor\`{u}b\'{a} topic.

\begin{table*}[h!]
\centering
\small
\begin{tabular}{l|l|l}
\hline
\textsc{Field} & \textsc{Description} & \textsc{Example}\\ \hline
1. Question ID & Unique identifier & 3506772758530306034 \\ \hline
2. English Document & English text document \\ \hline
3. English Question & Question in English & what is the name of the first nigerian \\& & president  \\ \hline
4. English Long Answer & Detailed answer in English &.ky is the Internet country code top-level\\ & &domain (ccTLD) for the Cayman [..] \\ \hline
5. English Short Answer & Brief answer in English & Nnamdi Azikiwe \\ \hline
6. Yor\`{u}b\'{a} Document & Yor\`{u}b\'{a} text document \\ \hline
7. Yor\`{u}b\'{a} Rewrite Flag & Was Yor\`{u}b\'{a} document rewritten? & 1 \\& (0: no, 1: yes) \\ \hline
8. Yor\`{u}b\'{a} Question & Question in Yor\`{u}b\'{a} & kí ni ky dúró fún ní erékùṣù cayman \\ \hline
9. Yor\`{u}b\'{a} Short Answer & Brief answer in Yor\`{u}b\'{a} & Nnamdi Azikiwe ni Aare \ \\ \hline
10. Yor\`{u}b\'{a} Long Answer & Detailed answer in Yor\`{u}b\'{a} & Nnamdi Azikiwe ti o je Gomina Agba \\ & & nigbana di Aare, ipo to je fun ayeye, [..] \\ \hline
11. Yor\`{u}b\'{a} Paragraph Info & Contextual information & P2 \\ \hline
12. Answer Alignment & Semantic equivalence & 1\\ 
&(0: not literal, 1: literal) \\ \hline
\end{tabular}
\caption{Dataset Fields, Descriptions and Sample entry.}
\label{tab:dataset-fields}
\end{table*}

\begin{table}[h!]
\centering
\small
\begin{tabular}{c|c|c|c|c}
\hline
&\textsc{Lan} & \multicolumn{1}{l|}{{\sc R-1}}  & \multicolumn{1}{l|}{{\sc R-2}} & \multicolumn{1}{l}{{\sc{R-L}}} \\ \hline
\textsc{GPT4o} & \textsc{Eng} & 0.39 & 0.23 & 0.30  \\
&\textsc{ Yor} & 0.34 & 0.19 & 0.27 \\\hline
\textsc{o1mini}& \textsc{Eng} & 0.45  &  0.22& 0.30  \\
& \textsc{Yor} & 0.30& 0.14 & 0.22 \\\hline
\textsc{LlaMA} & \textsc{Eng} & 0.31 & 0.18 & 0.23 \\
&\textsc{ Yor} &0.20 & 0.15 & 0.18  \\\hline
\end{tabular}
\caption{Results for 3 LLM in terms of Rouge computed for the entire set of questions. Human Score is computed on 358 questions.}
\label{tab:f1_scores_gpt4o}
\end{table}

%{\color{blue} [TODO JOY] Link to the English dataset completed}

\begin{comment}
\begin{table*}[htbp]
\centering
\small
\begin{tabular}{ll}
\toprule
\textbf{English Article} &  \\ 
\midrule
\textbf{English Question} & what is the name of the first nigerian president \\ 
\midrule
\textbf{English Short Answer} &  \\ 
\midrule
\textbf{English Long Answer} & . ky is the Internet country code top - level domain (ccTLD) for the Cayman Islands. Registration was limited to residents and registered companies in the Cayman Islands with a local address, but this restriction was removed in September 2015. The Cayman Islands also has the international three letter code, CYM, and has won a bid to be awarded the. cym domain in a future expansion of the top level domain space. \\ 
\midrule
\textbf{Yor\`{u}b\'{a} Article} &  \\
\midrule
\textbf{Yor\`{u}b\'{a} Question} & kí ni ky dúró fún ní erékùṣù cayman \\ 
\midrule
\textbf{Yor\`{u}b\'{a} Short Answer} & Nnamdi Azikiwe ni Aare \\ 
\midrule
\textbf{Yor\`{u}b\'{a} long Answer} & Nnamdi Azikiwe ti o je Gomina Agba nigbana di Aare, ipo to je fun ayeye, nigbati ti Abubakar Tafawa Balewa si di ipo Alakoso Agba mu. \\ 
\bottomrule
\end{tabular}

\caption{An Example of English and Yor\`{u}b\'{a} Entry}
\label{tab:project-overview}
\end{table*}
\end{comment}

\section{Experiments}
\label{sec:experiments}

\paragraph{Baselines} We evaluate our dataset with GPT-4o\footnote{gpt-4o version 2024-08-06} \citep{openai2024gpt4technicalreport}, o1-mini\footnote{o1-mini version 2024-09-12}, and LlaMA-3.1-8b \citep{dubey2024llama3herdmodels}, therevy covering both open and closed models, as well as models of different sizes. For each Y-NQ entry, we prompt the models with the following formatted instructions.

\begin{verbatim}
"""
Given the following passage and
a question,answer the question
in a single paragraph with
information found in the passage.

####
PASSAGE

{document}
####
QUESTION

{question}
####
ANSWER

"""
\end{verbatim}

\paragraph{Evaluation.} We evaluate the results by comparing the generated text and the reference long answer using several Rouge \citep{lin-2004-rouge} versions (Rouge-1, Rouge-2, Rouge-L).% and BERTScore \cite{DBLP:journals/corr/abs-1904-09675}.

%{\color{purple} TODO EDUARDO Can we add results based on length of the document?  }

\paragraph{Automatic metrics.} Table \ref{tab:f1_scores_gpt4o} reports the results showing that Yor\`{u}b\'{a} consistently performs worse than English (e.g., losing 0.4 in Rouge-1). However, the Yor\`{u}b\'{a} task is much easier because the documents are much shorter, which means that answering the question becomes an easier task. Even if we prompt the model to only answer based on the in-context document, we can not discard the idea that English may get better results due to using the internal knowledge from the model. 

\begin{figure}[ht!]
\center
     \includegraphics[width=7cm]{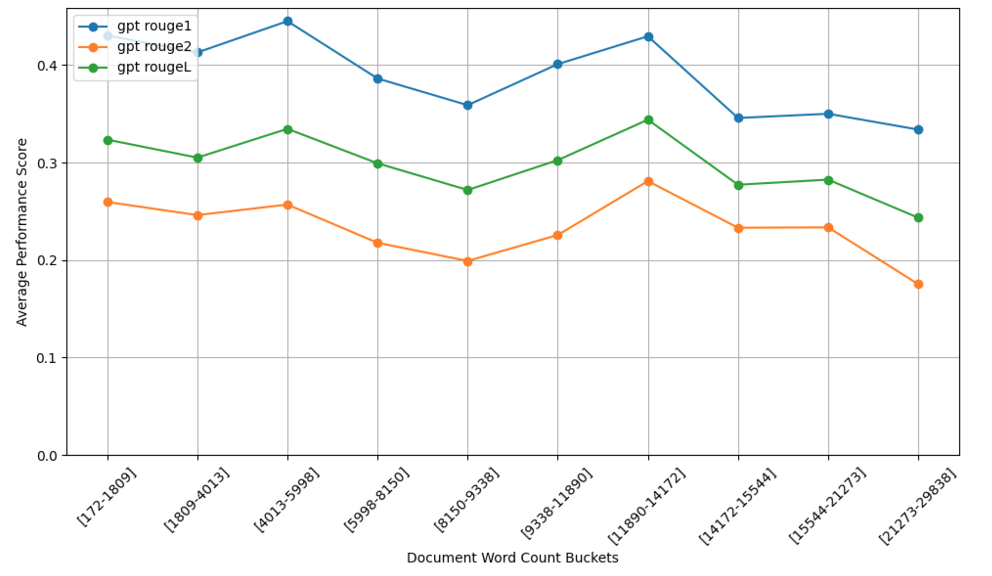}
    \includegraphics[width=7cm]{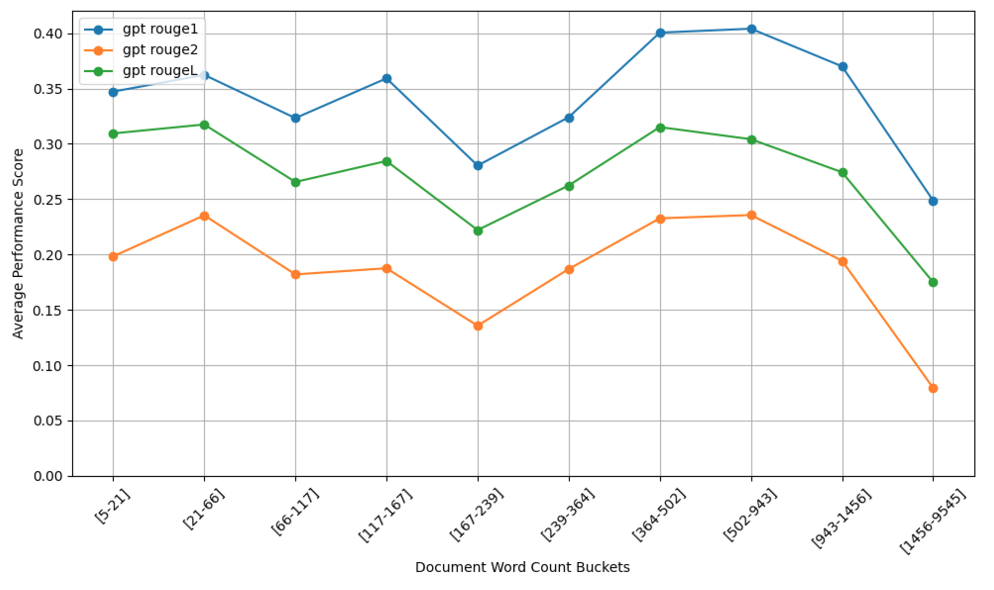}
    \caption{Impact of Document Length Buckets on Performance Scores for English (top) and Yor\`{u}b\'{a} (bottom) for GPT-4 outputs}
    \label{fig:doclength}
\end{figure}

\paragraph{Length analysis.} Model performance changes with the length of the document, as shown in Figure \ref{fig:doclength}. The dataset was split into equal size of documents in each length bucket. We can see a drop in performance when the Yor\`{u}b\'{a} documents reach 1,500 words, which shows the challenges that current models face in long-context understanding of low-resource languages. For a small portion of long-enough documents of comparable length between English and Yor\`{u}b\'{a} (only 4 documents that are over 900 words long), English performance demonstrates a significant edge (1.58X-2.56X), see Table \ref{tab:f1_scores_comparable}.

\begin{table}[h!]
\centering
\small
\begin{tabular}{l|l|l|l|l}
\hline
& \textsc{Avg W.} &  \textsc{R-1} &\textsc{R-2} & \textsc{R-L} \\ \hline
\textsc{Eng} & 3299 & 0.45 & 0.23 & 0.30  \\
\textsc{Yor} & 3070 & 0.32 & 0.09 & 0.19  \\
%doc count & 6 & 6 \\ 
\hline
\end{tabular}
\caption{Results for six comparable English and Yor\`{u}b\'{a} documents }%{\color{red} Joy can you add R1 and human eval for these docs? -> human eval scores are not calibrated, let's exclude them} }
\label{tab:f1_scores_comparable}
\end{table}

\section{Conclusions}

 Y-NQ is a newly released dataset that enables to compare generative open-book reading comprehension between English and Yor\`{u}b\'{a}. The main contributions of our data set are to allow for the comparison of LLM results in a reading comprehension task across a high- and a low-resource language, showing what are the generalization capabilities of LLMs in this particular case. Moreover, our annotations confirmed variations in the accuracy of Wikipedia articles in all languages. In particular, we identify inaccurate English responses for Yor\`{u}b\'{a} language-specific content. Y-NQ allows us to evaluate how reading comprehension capabilities extend to Yor\`{u}b\'{a}. Y-NQ is not exactly comparable in its totality between languages. Given that Yor\`{u}b\'{a} has shorter documents than English, the reading comprehension task is easier for Yor\`{u}b\'{a}. Therefore, results on this language should be much better than in English to expect parity between languages. Our experiments show that the reading comprehension capabilities of current English LLMs do not extend to Yor\`{u}b\'{a}.  Y-NQ is freely available on HuggingFace.%\footnote{{\color{red} ADD LINK}}.%As further work, we plan to perform human evaluation.

\section*{Limitations and Ethical considerations}

Y-NQ is limited in size, language, and domain coverage. The fact of using Wikipedia and extending an existing open-source dataset (NQ) may play in favor of having higher results in both languages due to contamination. Furthermore, the data set is not fully comparable between English and Yor\`{u}b\'{a}, since documents and answers vary in length. %Another challenge is that, due to the differences in the nature of the languages, the Likert scales in human evaluation are not exactly the same in English and Yor\`{u}b\'{a}. In Yor\`{u}b\'{a}, tone marks and spelling are more distinct features for judgment, while in English, annotator judgment relies more on spelling and word choice. The results are still comparable, but we should keep in mind that there is a discrepancy between the scales in the two languages. Our experimentation is limited to models and automatic evaluation metrics, which is compensated for through human evaluation.{\color{red} English references are not of high-quality?}

Our experimentation is limited to models and automatic evaluation metrics, which could be compensated for through human evaluation.
Annotators were paid a fair rate. 

\section*{Acknowledgements}

This paper is part of the LCM project\footnote{https://github.com/facebookresearch/large$\_$concept$\_$models} and the authors would like to thank the entire LCM team for the fruitful discussions. 

\bibliographystyle{plainnat}
\bibliography{custom}
\end{document}